\documentclass{article}

\PassOptionsToPackage{numbers, compress}{natbib}

\usepackage[final]{neurips_2019}
\usepackage[utf8]{inputenc} 
\usepackage[T1]{fontenc}    
\usepackage{url}            
\usepackage{booktabs}       
\usepackage{amsfonts}       
\usepackage{nicefrac}       
\usepackage{microtype}      
\usepackage{subcaption}
\usepackage{caption}
\usepackage{graphicx}
\usepackage{float}
\usepackage{mwe}
\usepackage{amsmath}
\usepackage{amssymb}
\usepackage{mathtools}
\usepackage[colorlinks=true,allcolors=blue]{hyperref}%
\usepackage{bm}

\title{Grassmannian Packings in Neural Networks: \\ Learning with Maximal Subspace Packings \\ for Diversity and Anti-Sparsity}

\author{%
  Dian Ang Yap \\
  Stanford University\\
  Stanford, CA 94305 \\
  \texttt{dayap@stanford.edu} \\
  \And
  Nicholas Roberts\\
  Carnegie Mellon University\\
  Pittsburgh, PA 15213\\
  \texttt{nick11roberts@cmu.edu} \\
  \And
  Vinay Uday Prabhu \\
  UnifyID AI Labs \\
  Redwood City, CA 94063 \\
  \texttt{vinay@unify.id} \\
}

\begin{document}

\maketitle

\begin{abstract}
Kernel sparsity (``dying ReLUs'') and lack of diversity are commonly observed in CNN kernels, which decreases model capacity. Drawing inspiration from information theory and wireless communications, we demonstrate the intersection of coding theory and deep learning through the Grassmannian subspace packing problem in CNNs. We propose Grassmannian packings for initial kernel layers to be initialized maximally far apart based on chordal or Fubini-Study distance. Convolutional kernels initialized with Grassmannian packings exhibit diverse features and obtain diverse representations. We show that Grassmannian packings, especially in the initial layers, address kernel sparsity and encourage diversity, while improving classification accuracy across shallow and deep CNNs with better convergence rates.
\end{abstract}

\section{Introduction}
\begin{minipage}{0.67\textwidth}
Filter level sparsity, commonly known as  ``dying kernels'' or ``dying ReLUs'' are common phenomena in Convolutional Neural Networks (CNNs). This phenomenon has been attributed to adaptive optimizers (Adam, Adagrad, Adadelta), high L2 or weight decay, and the usage of the ReLU activation function \citep{mehta2019implicit, lu2019dying, yaguchi2018adam}. Attempts to mitigate the dying ReLU issue with Leaky ReLUs have not been entirely effective, with minor overall impact on the emergent sparsity \citep{mehta2019implicit}.
\end{minipage}\hfill
\begin{minipage}{0.3\textwidth}
\centering
\begin{minipage}{.5\textwidth}
  \centering
  \includegraphics[width=.9\linewidth]{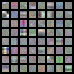}
  \label{fig:sparse_kernel}
\end{minipage}%
\begin{minipage}{.5\textwidth}
  \centering
  \includegraphics[width=.9\linewidth]{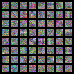}
  \label{fig:grass_kernel}
\end{minipage}
\captionof{figure}{\small \textit{Left}: sparse filters in CNNs; \textit{Right}: diverse, non-sparse Grassmannian kernels.}
\end{minipage}
\noindent

While kernel sparsity or pruning could benefit CNNs for faster inference and smaller models \citep{han2015deep, molchanov2016pruning, li2016pruning, gale2019state}, sparsity in the lower layers could be detrimental as the first few layers, especially the first layer, are ``sensitive to initialization'' are critical to forming good predictions \citep{zhang2019all}. Since the initial layers learn simple patterns and subsequent layers build upon initial layers to develop complex features, sparsity in initial layers would decrease model capacity.

Besides sparsity, kernel diversity has garnered much attention: Zeiler et. al. proposed a set of best practices to improve upon \textit{Alexnet} with different kernel sizes, stride-lengths and feature-scale clipping \citep{zeiler2014visualizing} to prevent specific filters from ``dominating''. Capturing distinctive features, especially in lower level layers, ensures that each filter learns something different and improves computational capacity that the CNNs dispense at the classification problem\citep{graff2016correlating}.

Here, we seek to address two problems, namely \textit{sparsity} and \textit{lack of diversity} of kernels, especially for the most important first few layers. We propose Grassmannian subspace packings as an initialization to project channel information onto a space where the basis vectors are maximally distant to capture as much diversity as possible and increase model capacity. We refer to the latter as ``channel expander'' layers. We demonstrate that models with first-layer Grassmannian kernels achieve better accuracy and convergence rate with diverse, non-sparse kernels, which improves model capacity.

\section{Grassmannian line and subspace packings}
Grassmannian subspace packing \citep{conway1996packing} in experimental mathematics and information theory has been used in beamforming techniques for physical layer wireless communications, limited feedback codebook-based downlink beamforming schemes and sparse signal reconstruction \citep{malioutov2005sparse, love_1, prabhu2009performance}.

\textbf{Set theoretic definition.} The real Grassmann manifold $G(m,k)$ is define as the set of all $k$-dimensional subspaces in $\mathbb{R}^m$, where $G(m,k) = \{W \subset \mathbb{R}^m \; | \; \text{dim}(W) = k\}$.
 The Grassmannian N-subspace packing problem is the problem of finding a set of $N$ $k$-dimensional subspaces in $G(m,k)$ that maximize the minimum pairwise distance between subspaces in the set under some metric \citep{dhillon2008constructing}. We define the pairwise distance between any two subspaces in $\mathbb{R}^m$ in terms of principal angles.

\textbf{Principle angles.} Suppose that $\mathcal{S}$ and $\mathcal{T}$ are two subspaces in $G(m, k)$. These subspaces are inclined against one another by $k$ different principle angles, with the smallest principal angle $\theta_1$ formed by a pair of unit-length vectors $(s_1, t_1)$ drawn from $\mathcal{S} \times \mathcal{T}$, where

\begin{equation}
    \theta_1 = \inf_{(s_1, t_1) \in \mathcal{S} \times \mathcal{T}} \arccos \langle s_1, t_1\rangle \quad \quad \text{ where $||s_1||_2 = 1$ and  $||t_1||_2 = 1$}
\end{equation}

with $\langle x, y \rangle$ denoting the inner product of vectors $x$ and $y$, and $||x||_2$ denoting the $\ell_2$ norm metric such that $||x||_2 = \langle x, x\rangle$.

The second principle angle $\theta_2$ is defined as the smallest angle obtained by a pair of unit length vectors $(s_2, t_2)$ from the same subspace $\mathcal{S} \times \mathcal{T}$ that is orthogonal to the first pair, where

\noindent\begin{minipage}{.55\linewidth}
\begin{equation*}
  \theta_2 = \inf_{(s_2, t21) \in \mathcal{S} \times \mathcal{T}} \arccos \langle s_2, t_2\rangle
  \end{equation*}
\end{minipage}%
\begin{minipage}{.45\linewidth}
\begin{equation}
  \begin{split}
    \text{where   } & ||s_2||_2 = 1 \; , \; ||t_2||_2 = 1\\
    \text{and } & \langle s_1, s_2\rangle = 0 \;, \; \langle t_1, t_2\rangle = 0.
      \end{split}
  \end{equation}
\end{minipage}
\vspace*{0.1cm}

The remaining principle angles $\theta_3, \dots, \theta_k$ are defined similarly. The sequence of principle angles $\{ \theta_1, \dots, \theta_k\}$ is non-decreasing and bounded in $[0, \pi/2]$.

\textbf{Optimizing Grassmannians using principle angles.} For an $N$ packing problem of $G(m, k)$, let $\Omega_{m\times k}$ denote the set of subspaces in $\mathbb{R}^{m\times k}$. We find $\Omega_{m \times k}$  where the minimum distance between any $\mathbf{w}_i, \mathbf{w}_j \in \Omega_{m \times k} (i \neq j)$ is maximized. The two popular distance metrics being chordal distance and Fubini-Study distance, which are defined below. With the restriction that $k \leq m$, for two $k$-dimensional subspaces $\mathcal{S}$ and  $\mathcal{T}$,

\noindent\begin{minipage}{.5\linewidth}
\begin{equation}
  \begin{split}
    d_{\text{chord}}(\bm{\mathcal{S}}, \bm{\mathcal{T}})&\coloneqq \sqrt{\sin^2\theta_1 + \dots + \sin^2 \theta_k} \\
    &= \left[k - ||\bm{S} * \bm{T}||_F^2 \right]^{\frac{1}{2}}.
      \end{split}
  \end{equation}
\end{minipage}%
\begin{minipage}{.5\linewidth}
\begin{equation}
  \begin{split}
    d_{\text{FS}}(\bm{\mathcal{S}}, \bm{\mathcal{T}}) &\coloneqq \arccos \left( \prod_k \cos \theta_k' \right)\\
    &= \arccos|\det \bm{S}*\bm{T} \;|.
      \end{split}
  \end{equation}
\end{minipage}
\vspace*{0.1cm}

For our experiments, we use generate a codebook matrix for subspace packing based on Fubini-Study distance which has better experimental results\footnote{\url{https://www.mathworks.com/matlabcentral/fileexchange/41652-grassmannian-design-package}.}\citep{medra2014flexible}. We first generate a \textit{codebook matrix}, $\mathbf{W}=\{\mathbf{w}_1 | \mathbf{w}_2| ... |\mathbf{w}_N\};\mathbf{w}_i \in \Omega_{m \times k}$. $\mathbf{W}$ is a rank-$k$ codebook characterized by the minimum distance of packing $\delta(\mathbf{W})$,

\begin{equation}
    \delta(\mathbf{W}) \coloneqq \min_{i \neq j} d(\mathbf{w}_i, \mathbf{w}_j)
\end{equation}

We seek to find $\mathbf{W}=\{\mathbf{w}_i\}_{i=1}^N$ that maximizes $\delta(\mathbf{W})$,

\begin{equation}
    \max_{\{\mathbf{w}_i\}, \mathbf{w}_i \in G(m, k)} \min_{i \neq j} \; d(\mathbf{w}_i, \mathbf{w}_j)
\end{equation}

The Rankin bound \citep{barg2002bounds} provides the theoretical upper bound of the maximum of the minimum pairwise distances between subspaces and is given by,

\begin{equation}
    \delta ({\mathbf{W}}) \leq \sqrt {\frac{{(m - k)N}}{{m(N - 1)}}}
\end{equation}

Grassmannian packing seeks to answer \textit{"what is the best way to optimally pack N k-dim subspaces in an M-dim space?"}. For the first convolutional kernel with width = 3, height = 3 (9 parameters per channel) with input channels = 3, output channels = 32, we ask:

\begin{center}
\textit{"How can we optimally pack 9 3-dim subspace in a 32-dim space, \\ such that any two subspaces (kernels) are maximally separated?"}
\end{center}
\section{Experiments}
We first generate Grassmannian codebooks for subspace packings, and load the codebook as CNN kernels. We compare accuracy and rate of convergence of shallow CNNs and deep CNNs. For a shallow 4-conv CNN, we trained 3 models: Kaiming-initialized CNNs, the same CNN but with Grassmannian as first layer (frozen and trainable). These models had their first layer initialized as 32 packings of $G(9, 3)$ for CIFAR10 and CIFAR100, and 32 packings of $G(9, 1)$ for single-channel MNIST and KMNIST\footnote{The first conv layer has parameters (width 3, height 3, 1 or 3 input channels and 32 output channels).}.

For deep CNNs, we used ResNet18 architecture and initialized the first layer with 64 packings of $G(49, 3)$\footnote{The first layer of ResNets are conv kernels (width 7, height 7, 3 input channels and 64 output channels).}. We trained ResNet with batch size of 64, SGD optimizer with learning rate of 0.01, momentum of 0.9, with a learning rate decay by a factor of $\sqrt{10}$ every 10 epochs \citep{he2016deep}. Besides having Kaiming-initialized baseline, first-layer Grassmann and first-layer frozen Grassmann, we also scaled the Grassmann as per Kaiming init, such that the activations are correctly scaled that facilitates learning without vanishing gradients \citep{he2015delving}. We scale the Grassmannian basis such that the magnitude is of factor $\sqrt{2/d_{\text{in}}}$. Since the scaling factor is constant, the directions of Grassmannian subspace basis vectors are preserved.

We check for convergence rate by examining first-epoch accuracy of shallow CNNs on MNIST, KMNIST, CIFAR10 and CIFAR100, as well as accuracy, kernel sparsity and diversity. As adaptive optimizers and weight decay could affect sparsity, we test on different optimizers such as SGD, Adadelta and Adam, and with/without weight decay of 0.0001.

\section{Results and Discussions}
Grassmannian models achieve higher accuracy on the 10 class Imagenette \cite{imagenette} problem with ResNets, even when Grassmannian kernels are frozen upon initialization. With weight decay, Grassmannian kernels achieve the highest accuracy out of the four, whereas Kaiming-scaled Grassmannian packings outperforms the rest without weight decay, as per Table \ref{table:deep-results}.
\begin{table}[!htb]
  \caption{ResNet18 on 10-class ImageNette with SGD optimizer, with and without weight decay. }
  \label{table:deep-results}
  \centering
  \begin{tabular}{lcc}
    \toprule
     & \multicolumn{2}{c}{Validation Accuracy (\%)} \\
     \cmidrule(r){2-3}
    Model  &Without Decay & With Decay   \\
    \midrule
    Baseline                     & 90.7          & 90.4 \\
    Grassmannian                 & 92.4          & \textbf{91.8} \\
    Grassmannian, Frozen         & 90.8          & 91.6 \\
    Grassmannian, Kaiming-Scaled & \textbf{92.8} & 91.5 \\
    \bottomrule
    \end{tabular}
\end{table}

\begin{figure}[!ht]
  \centering
   \begin{tabular}{c}
   \includegraphics[width=0.33\textwidth]{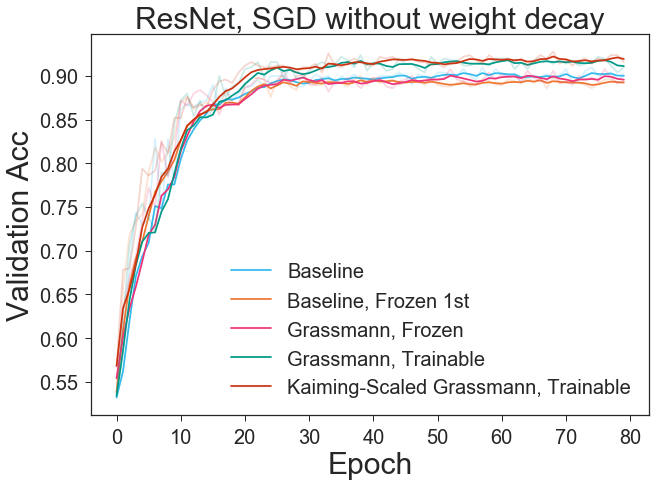}
   \includegraphics[width=0.33\textwidth]{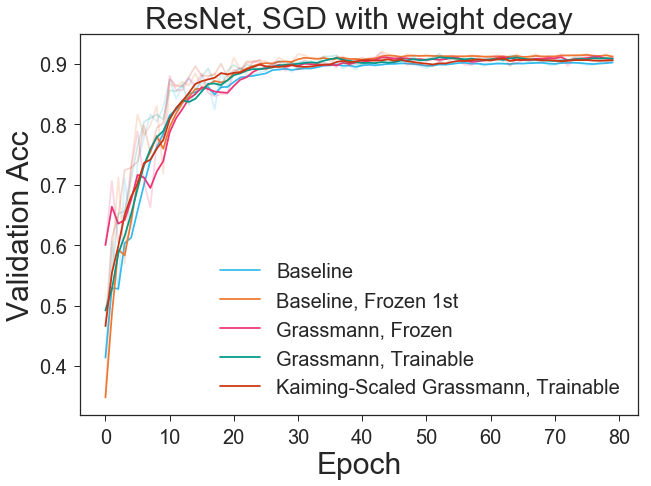}
   \includegraphics[width=0.33\textwidth]{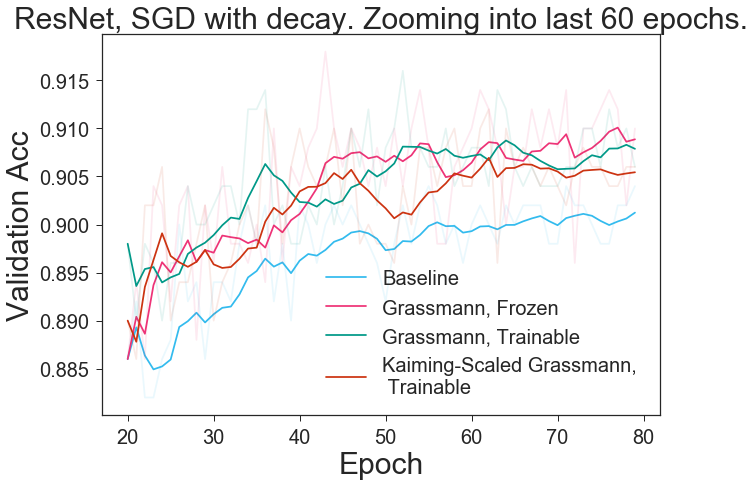}
   \end{tabular}
  \caption{ResNet with and without weight decay, with Grassmannian achieving better convergence and better final accuracy with everything else kept constant. Last image: zooming into last few epochs for ResNet with weight decay.}
   \label{fig:resnet}
\end{figure}

Grassmannian initialization also improves convergence for both shallow (Figure \ref{fig:shallow_convergence}) and Deep CNNs (Figure \ref{fig:resnet}), with both frozen and trainable Grassmannian models achieving higher first-epoch accuracy (Table \ref{table:shallow-results}). Grassmannian subspace initialization, where each convolutional filter was assigned the subspace basis vectors for their corresponding Grassmannian packing scheme, outperformed the baseline on every dataset.
\begin{table}[!htb]
  \caption{Faster Convergence of Shallow CNN, first epoch accuracy.}
  \label{table:shallow-results}
  \centering
  \begin{tabular}{lcccc}
    \toprule
     & \multicolumn{4}{c}{Validation Accuracy (\%)} \\
     \cmidrule(r){2-5}
    Model  & MNIST & KMNIST & CIFAR10 & CIFAR100  \\
    \midrule
    Baseline             & 90.9 & 59.7 & 33.6 & 2.57 \\
    Grassmannian, Frozen         & 91.9 & 66.2 & 40.5 & \textbf{4.73}  \\
    Grassmannian & \textbf{92.6} & \textbf{66.4} & \textbf{41.1} & 4.70 \\
    \bottomrule
    \end{tabular}
\end{table}

Interestingly, we see a marginal improvement over Grassmannian initialization by simply rendering the Grassmannian initialization untrainable. We postulate that in certain cases, simply projecting the input feature maps to be as far apart as possible as opposed to using another trainable convolutional layer is enough for later layers to make better use of their own capacity, at least in early training.

\begin{figure}[!ht]
  \centering
   \begin{tabular}{c}
   \includegraphics[width=0.33\textwidth]{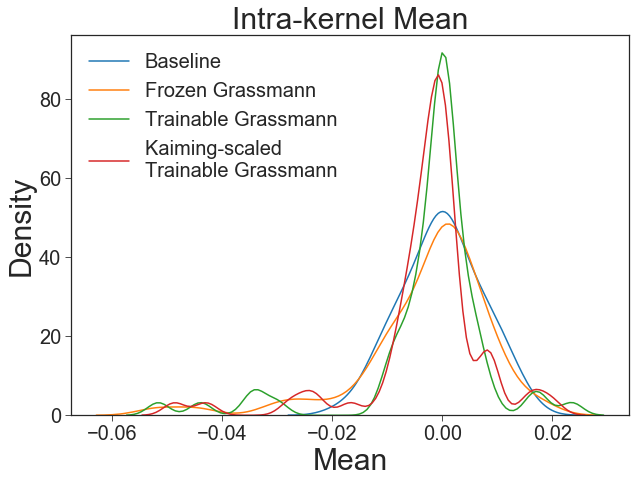}
   \includegraphics[width=0.33\textwidth]{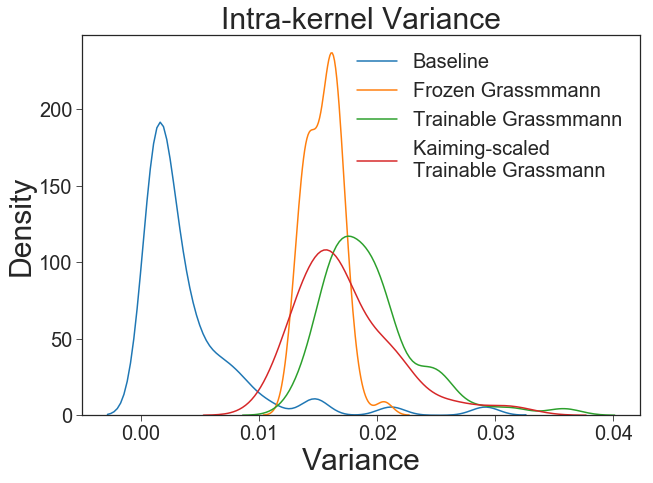}
   \includegraphics[width=0.33\textwidth]{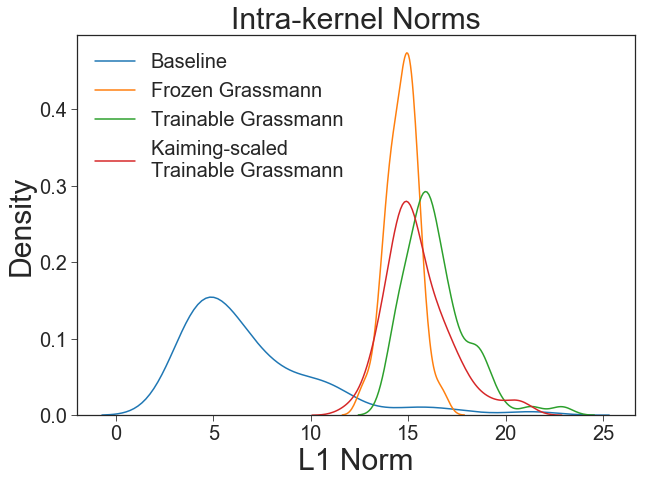}
   \end{tabular}
  \caption{Distribution of mean, mariance and norm over the 64 kernels of first layer of ResNet.}
   \label{fig:mean_var_norm}
\end{figure}

\begin{figure}[!ht]
    \centering
    \begin{subfigure}[b]{0.475\textwidth}
        \centering
        \includegraphics[width=5cm]{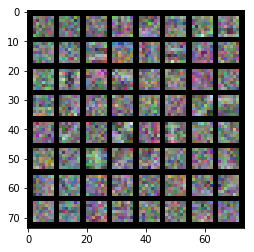}
        \caption[]%
        {{\small Grassmannian, no weight decay.}}
    \end{subfigure}
    \hfill
    \begin{subfigure}[b]{0.475\textwidth}
        \centering
        \includegraphics[width=5cm]{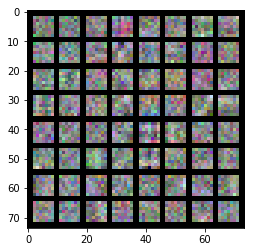}
        \caption[]%
        {{\small Kaiming-scaled  Grassmannian, no weight decay.}}
    \end{subfigure}
    \vskip\baselineskip
    \begin{subfigure}[b]{0.475\textwidth}
        \centering
        \includegraphics[width=5cm]{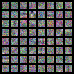}
        \caption[]%
        {{\small Grassmannian, weight decay.}}
    \end{subfigure}
    \quad
    \begin{subfigure}[b]{0.475\textwidth}
        \centering
        \includegraphics[width=5cm]{imgs/scale_kernel.png}
        \caption[]%
        {{\small Kaiming-scaled Grassmannian, weight decay.}}
    \end{subfigure}
    \caption
    {Visualizations of Trainable Grassmannian Kernels of ResNet trained using SGD, with and without weight decay after training for 80 epochs on ImageNette.}
    \label{fig:more_grass_kernels}
\end{figure}

Qualitative inspection of the kernels (Figure \ref{fig:more_grass_kernels}) reveal that Grassmannian kernels learn diverse features with a mixture of edge, color and pattern detectors in each kernel, without any kernel dominating that shows improved diversity and fewer sparsity. While baseline and Grassmannian packings gave mean close to 0, intra-kernel variance and norms are higher for Grassmannian packings, indicating diversity and fewer sparsity with the distribution across kernels shown in Figure \ref{fig:mean_var_norm}.

Interestingly, trainable Grassmannian packings outperform baseline which outperforms frozen Grassmannian packings when adaptive optimizers are used (especially Adam, see Figure \ref{fig:optimmatters}). Since Adam contributes to sparsity \citep{mehta2019implicit, yaguchi2018adam}, it is at odds with Grassmannian packings that are against sparsity and for diversity. Nonetheless, most models such as ResNet, RandWire, NASNet and DenseNet achieve SOTA results through SGD with a careful learning rate scheduler, which could further benefit from Grassmannian packings due to their demonstrated performance using SGD \citep{wilson2017marginal}.

\section{Conclusion}

In this work, we showed that the use of Grassmannian packings for first layer initialization improved performance in the contexts of shallow networks and deep networks, even with untrainable Grassmannian kernels, across multiple datasets. Grassmannian kernels are best used with SGD optimization and works with weight decay or otherwise. Our results suggest improvement of the model's capacity to learn by simply initializing the first layer filters to be maximally distant and capture as many diverse features as possible. In the future, we hope to extend this analysis to explore the several other options for metric used in the Grassmannian packing, and search for ways to inform the choice in metric.

\subsubsection*{Acknowledgments}
\begin{figure}[!ht]
    \begin{subfigure}[b]{.33\linewidth}
    \centering
    \includegraphics[width=\linewidth]{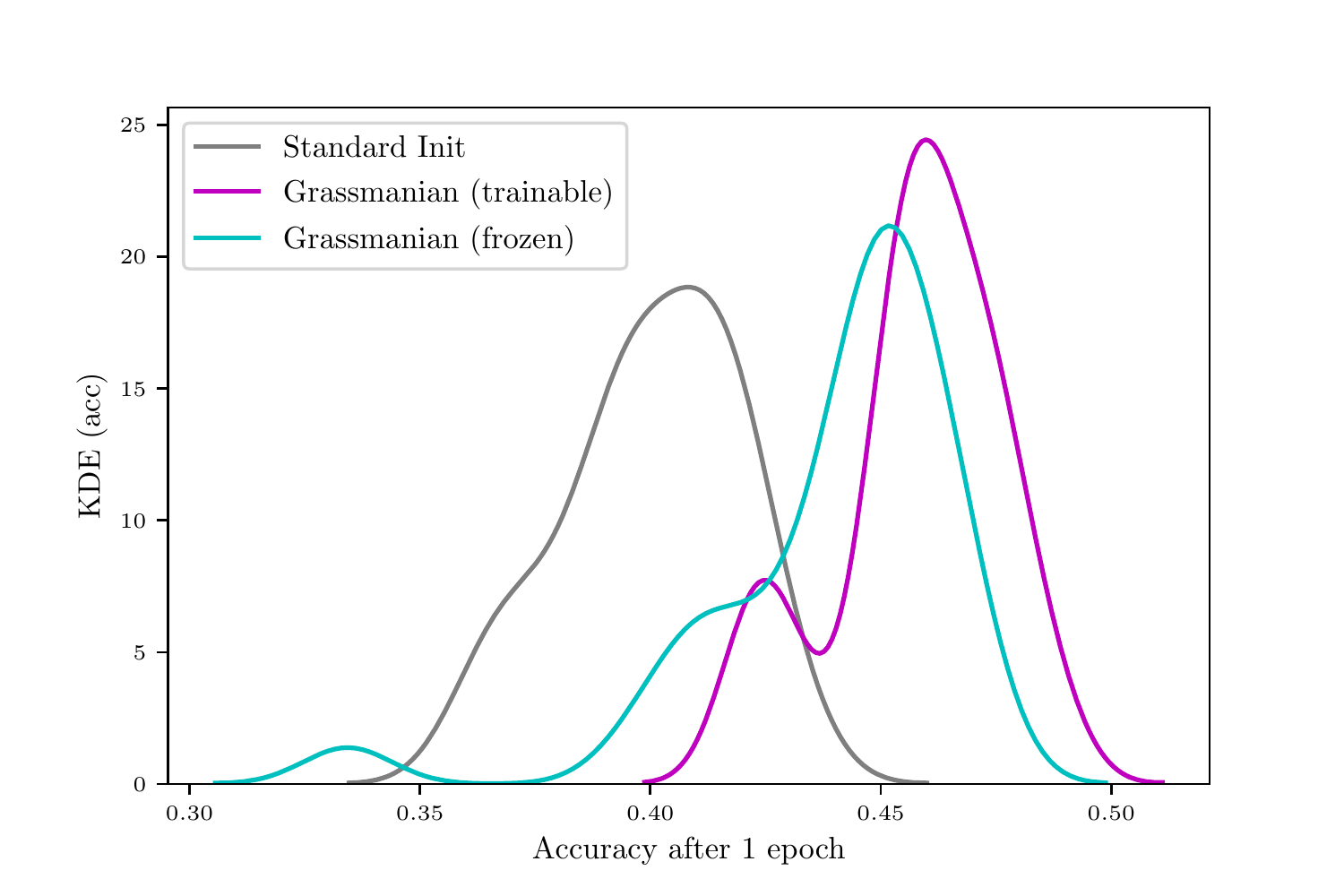}
    \caption{SGD}
    \end{subfigure}
    \begin{subfigure}[b]{.33\linewidth}
    \centering
    \includegraphics[width=\linewidth]{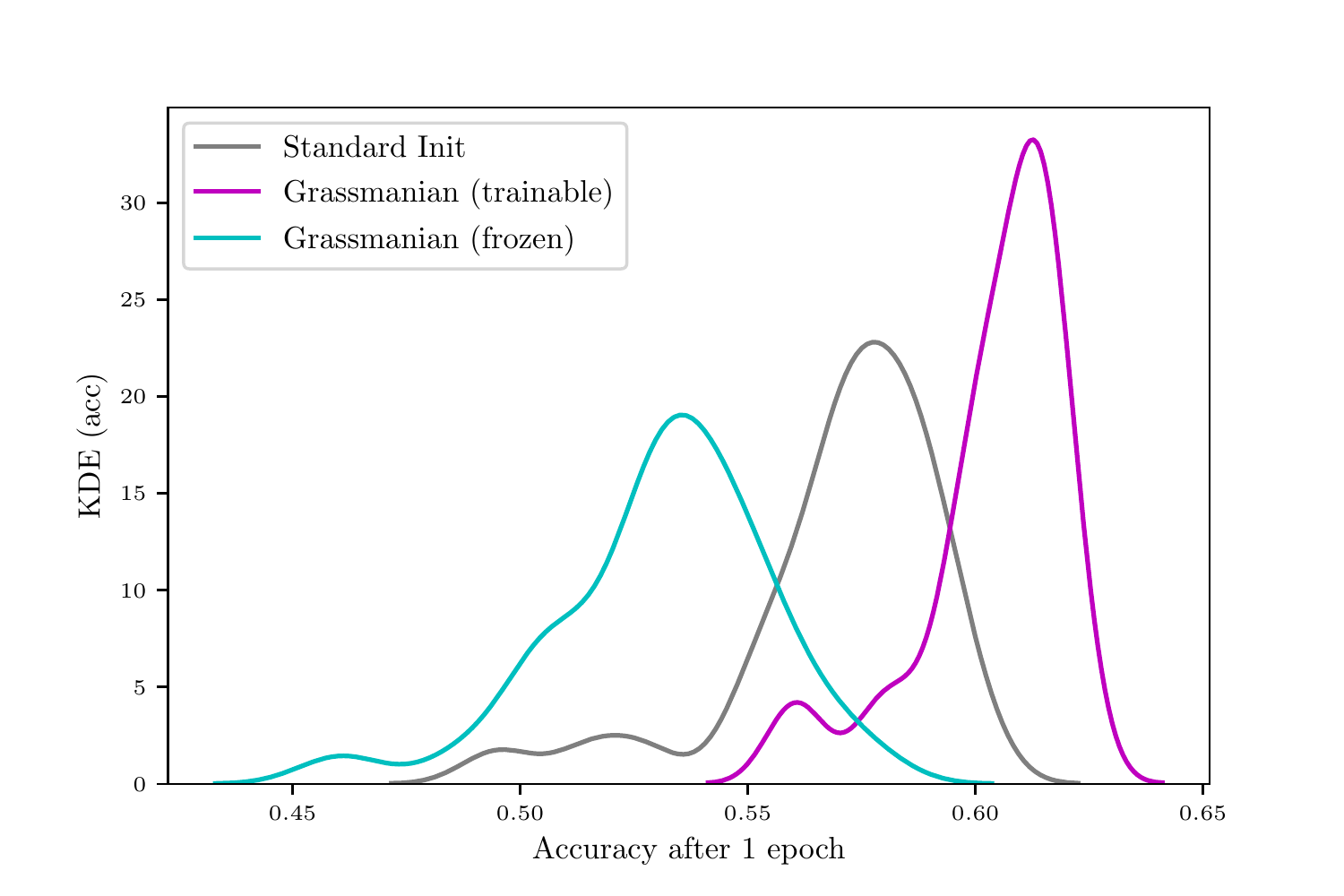}
    \caption{Adadelta}
    \end{subfigure}
    \begin{subfigure}[b]{.33\linewidth}
    \centering
    \includegraphics[width=\linewidth]{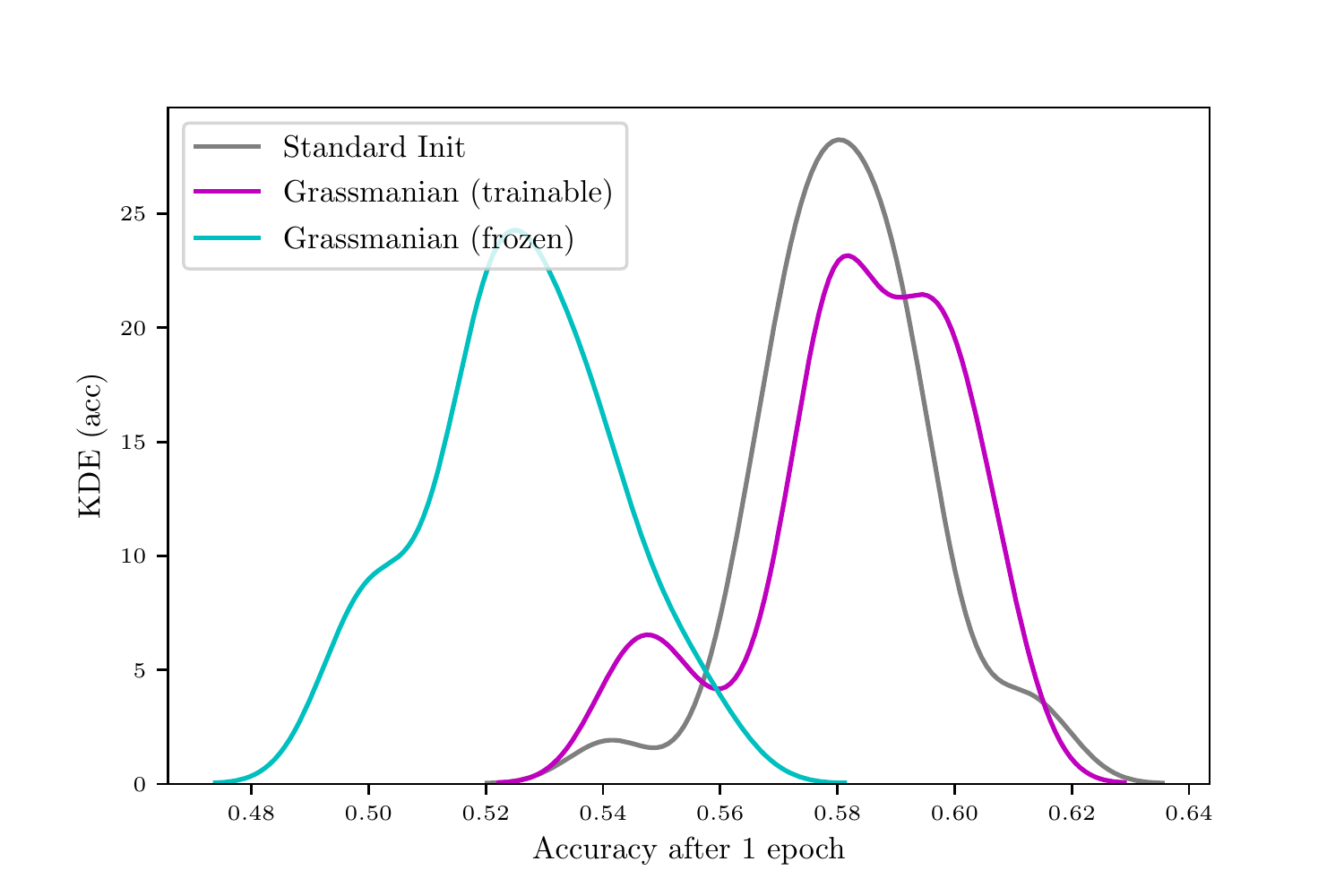}
    \caption{Adam}
    \end{subfigure}
    \caption{Distributions of 30 runs of first-epoch test accuracy at the first epoch \textbf{\textit{with different optimizers}} on the same shallow CNN architecture, comparing initialization of first layer using standard Xavier initialization, frozen Grassmannian packings, and trainable Grassmannian packings.}
    \label{fig:optimmatters}
\end{figure}

\begin{figure}[!ht]
    \centering
    \begin{subfigure}[b]{0.475\textwidth}
        \centering
        \includegraphics[width=\textwidth]{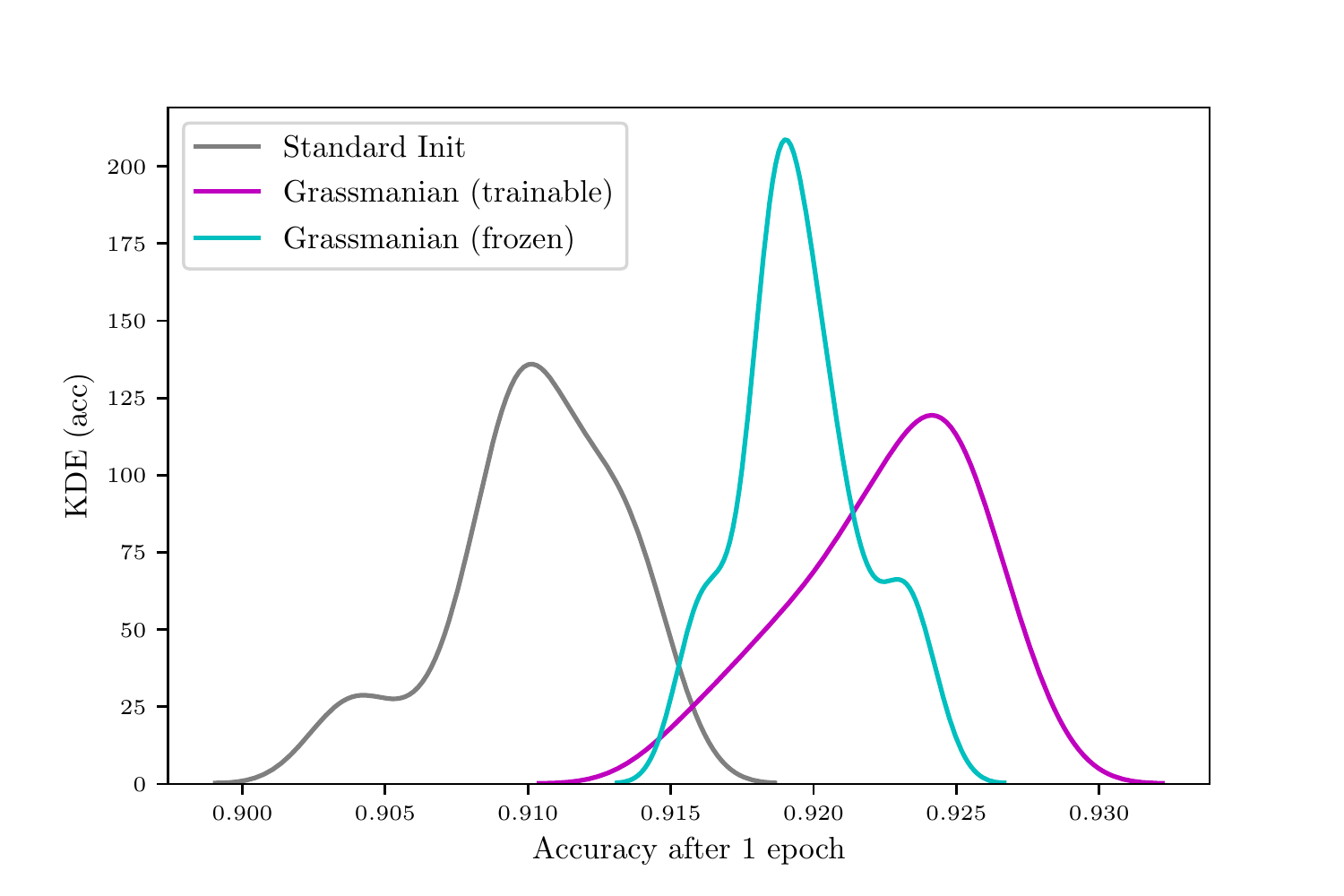}
        \caption[]%
        {{\small MNIST}}
    \end{subfigure}
    \hfill
    \begin{subfigure}[b]{0.475\textwidth}
        \centering
        \includegraphics[width=\textwidth]{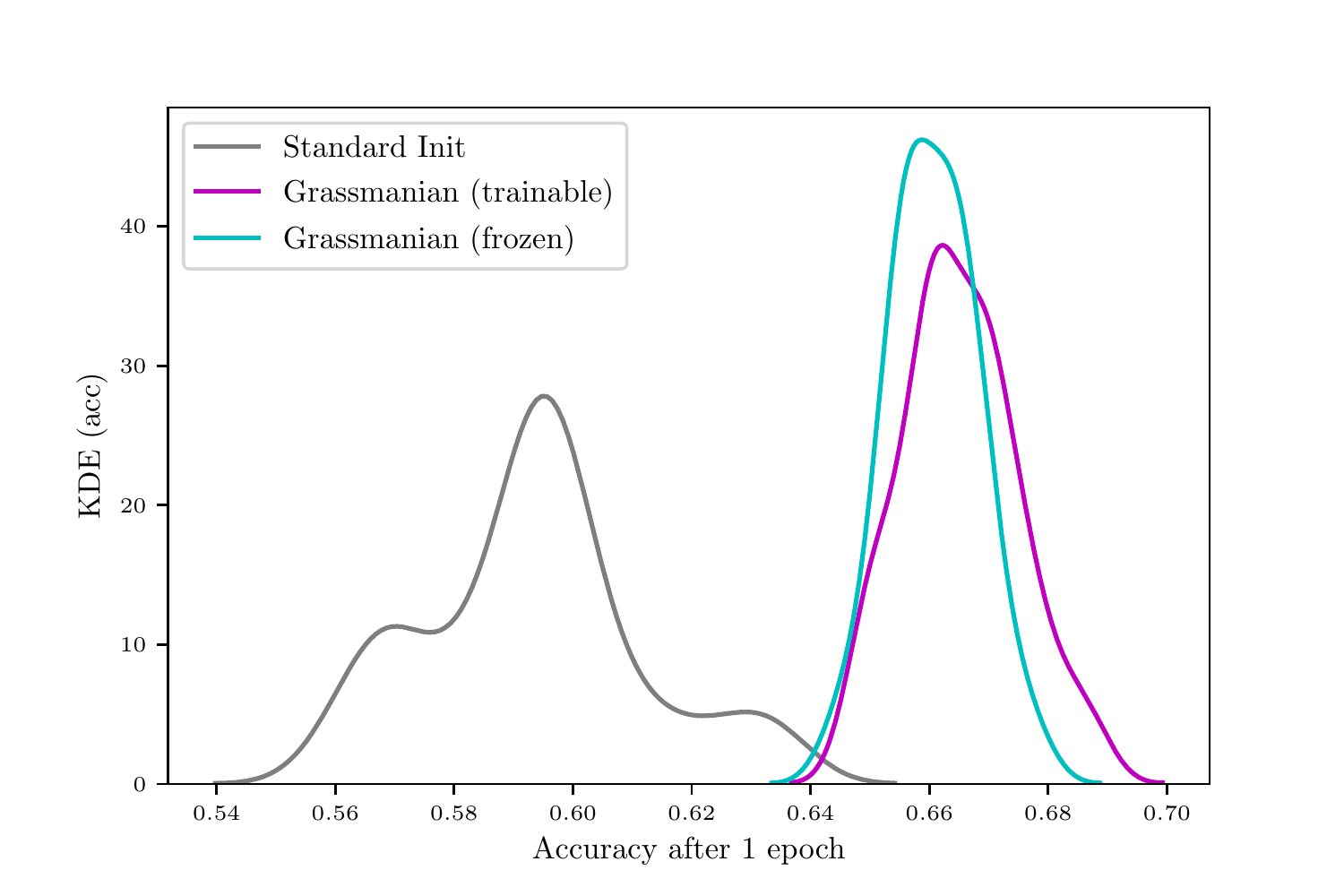}
        \caption[]%
        {{\small KMNIST}}
    \end{subfigure}
    \vskip\baselineskip
    \begin{subfigure}[b]{0.475\textwidth}
        \centering
        \includegraphics[width=\textwidth]{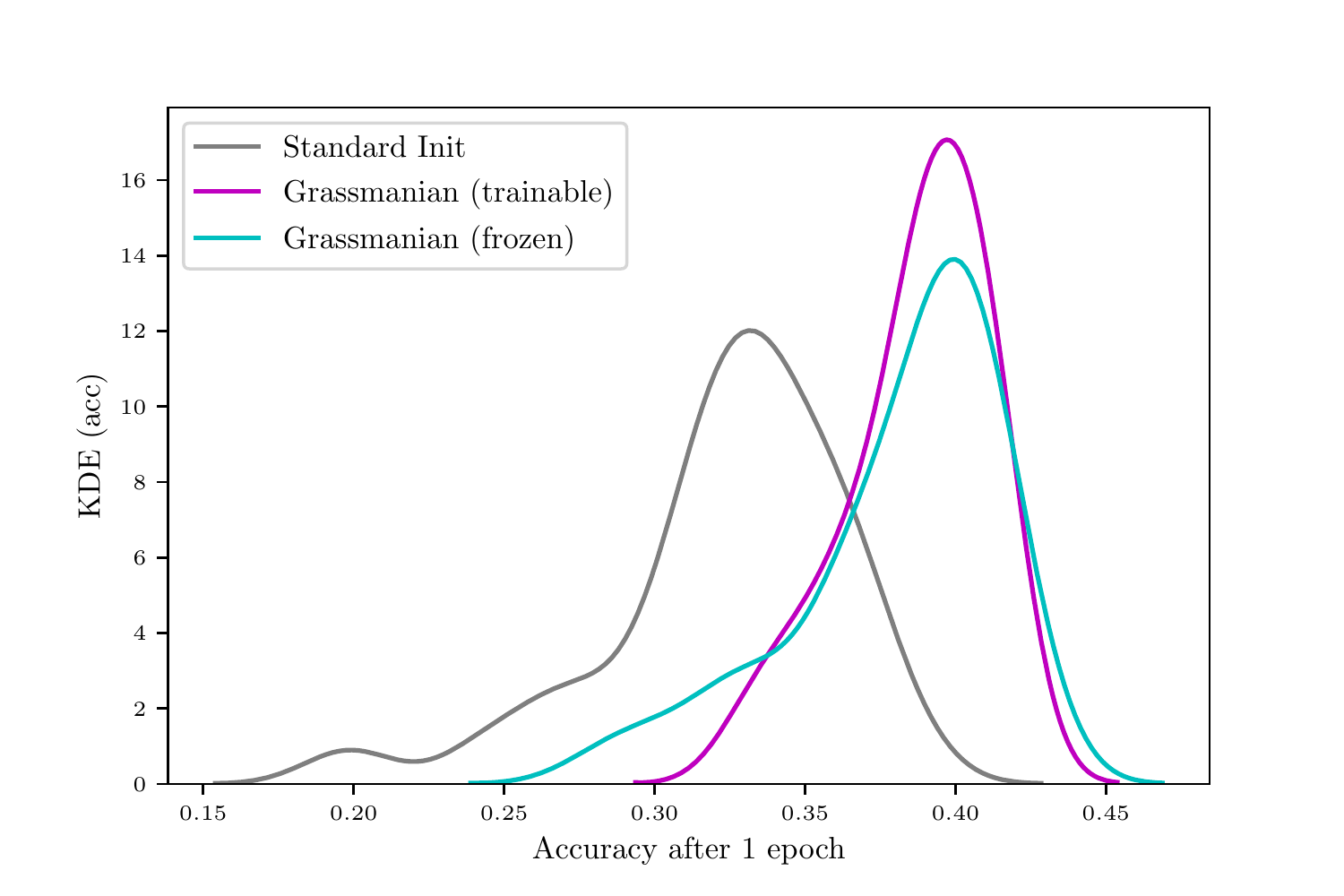}
        \caption[]%
        {{\small CIFAR10}}
    \end{subfigure}
    \quad
    \begin{subfigure}[b]{0.475\textwidth}
        \centering
        \includegraphics[width=\textwidth]{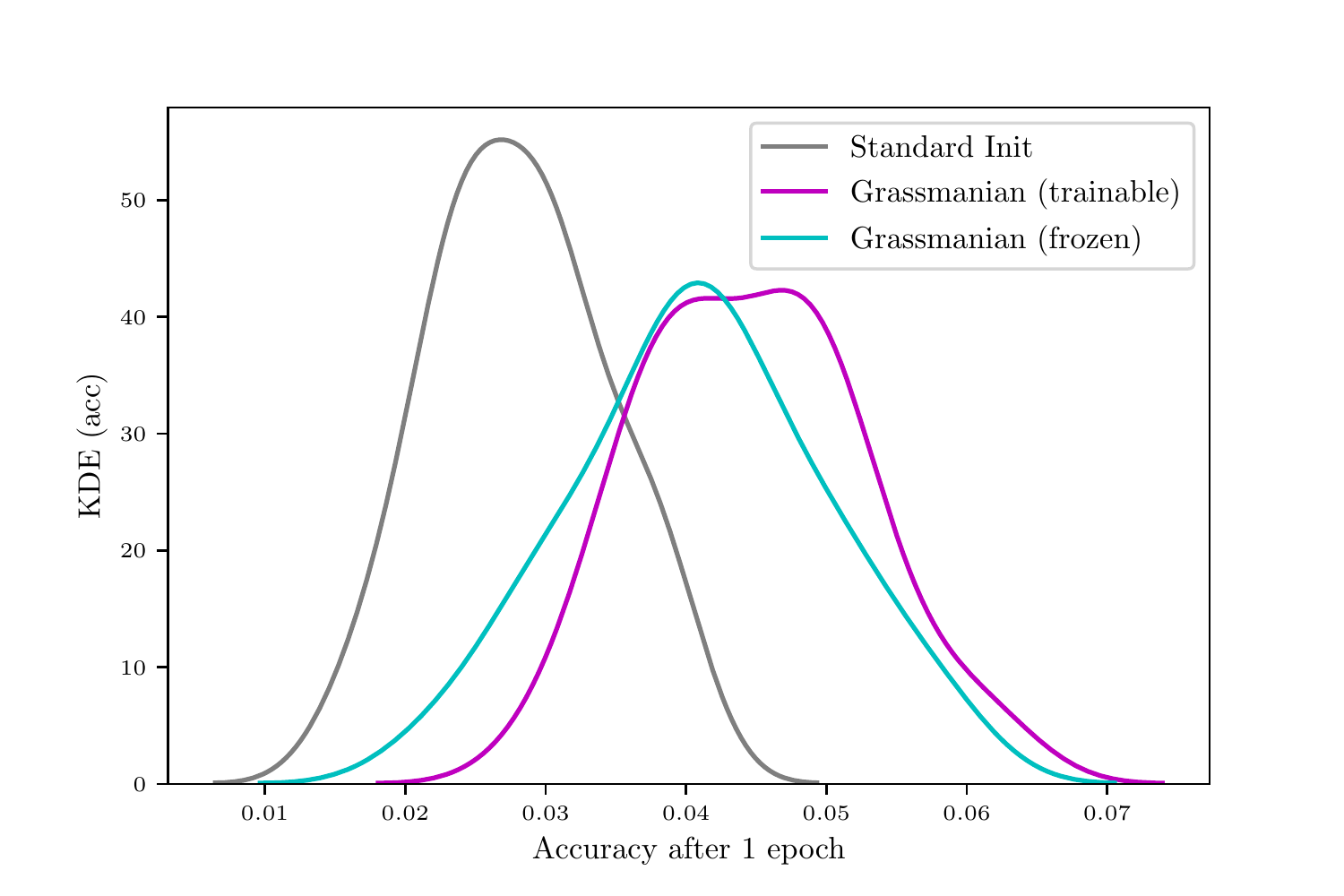}
        \caption[]%
        {{\small CIFAR100}}
    \end{subfigure}
    \caption
    {Distributions of 30 runs of first-epoch test accuracy at the first epoch with SGD with different datasets, comparing initialization of first layer using standard Xavier initialization, frozen Grassmannian packings, and trainable Grassmannian packings.}
    \label{fig:shallow_convergence}
\end{figure}

We would like to acknowledge Matthew McAteer for helpful discussions and contrbution of designs and plots for Figures \ref{fig:optimmatters} and \ref{fig:shallow_convergence}.

\small
\bibliography{ref}
\bibliographystyle{icml2019}

\section*{Appendix}

\begin{figure}[!htb]
    \vskip 0.1in
    \centering
    \includegraphics[width=\textwidth]{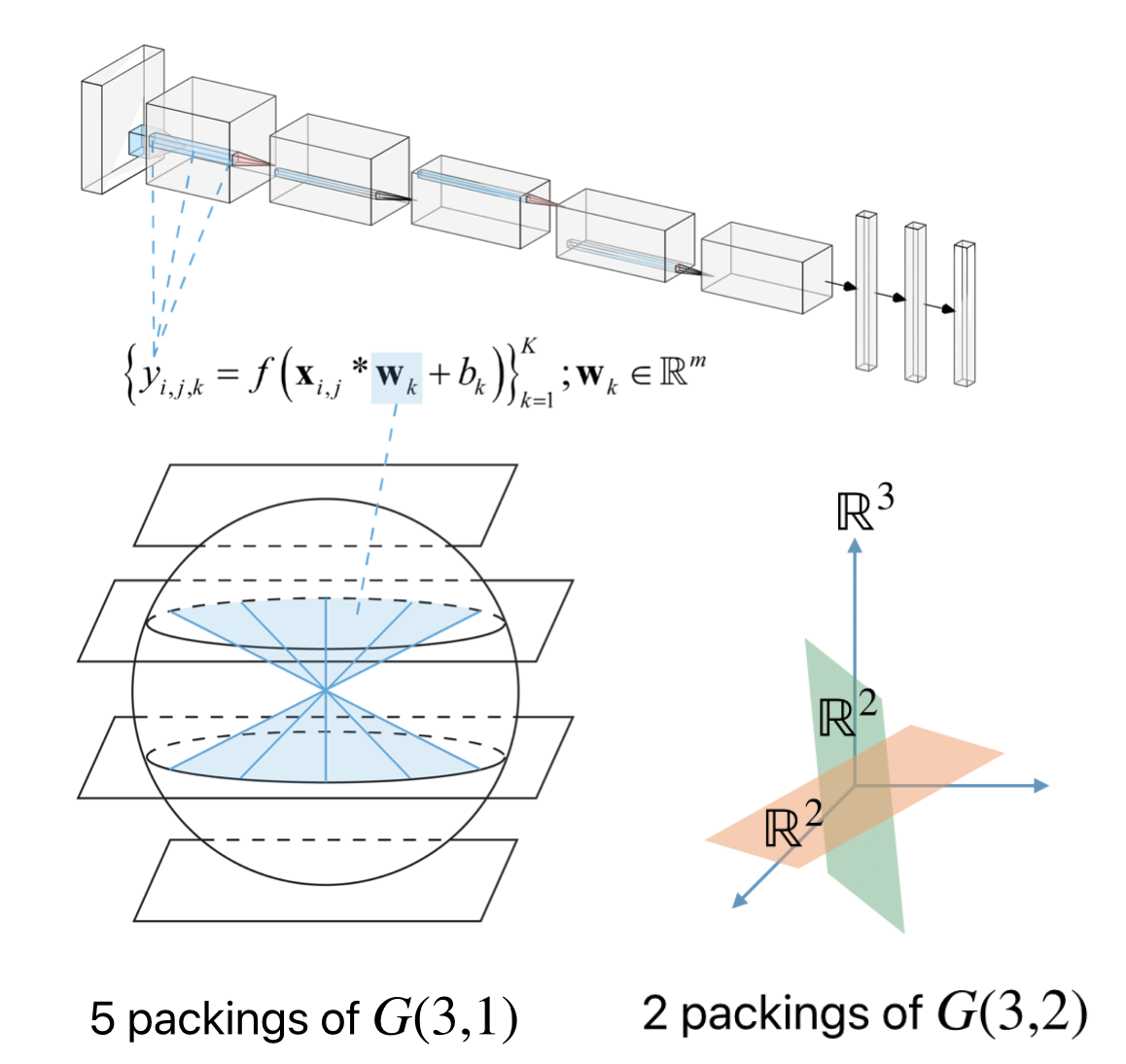}
    \caption{Illustration of how the first conv layer is initialized with Grassmannians, with visualizations of Grassmannian packings. \textit{Left}: 5-packing of $G(3, 1)$, which answers ``how should 5 laser beams passing through a single point in $\mathbb{R}^3$ be arranged so as to make the minimum principle angle between any two of the beams as large as possible". \textit{Right}: 2-packing of $G(3, 2)$. These concepts can be extended to higher dimensions.}
    \label{fig:resnet56}
\end{figure}

\begin{table}[!ht]
\caption{Final Accuracy of ResNet-56 on CIFAR-10 with first layer Xavier, Grassmannian trainable, and Grassmannian Frozen.}
\begin{sc}
\begin{center}
\begin{tabular}{cc}
\hline
Initialization & Test Acc\\
\hline
Standard, Xavier & 91.85 \\
Grassmannian, Fixed & 91.98\\
Grassmannian, Trainable & \textbf{92.31}\\
\hline
\end{tabular}
\end{center}
\end{sc}
\end{table}

\begin{figure}[!htb]
    \vskip 0.1in
    \centering
    \includegraphics[width=\textwidth]{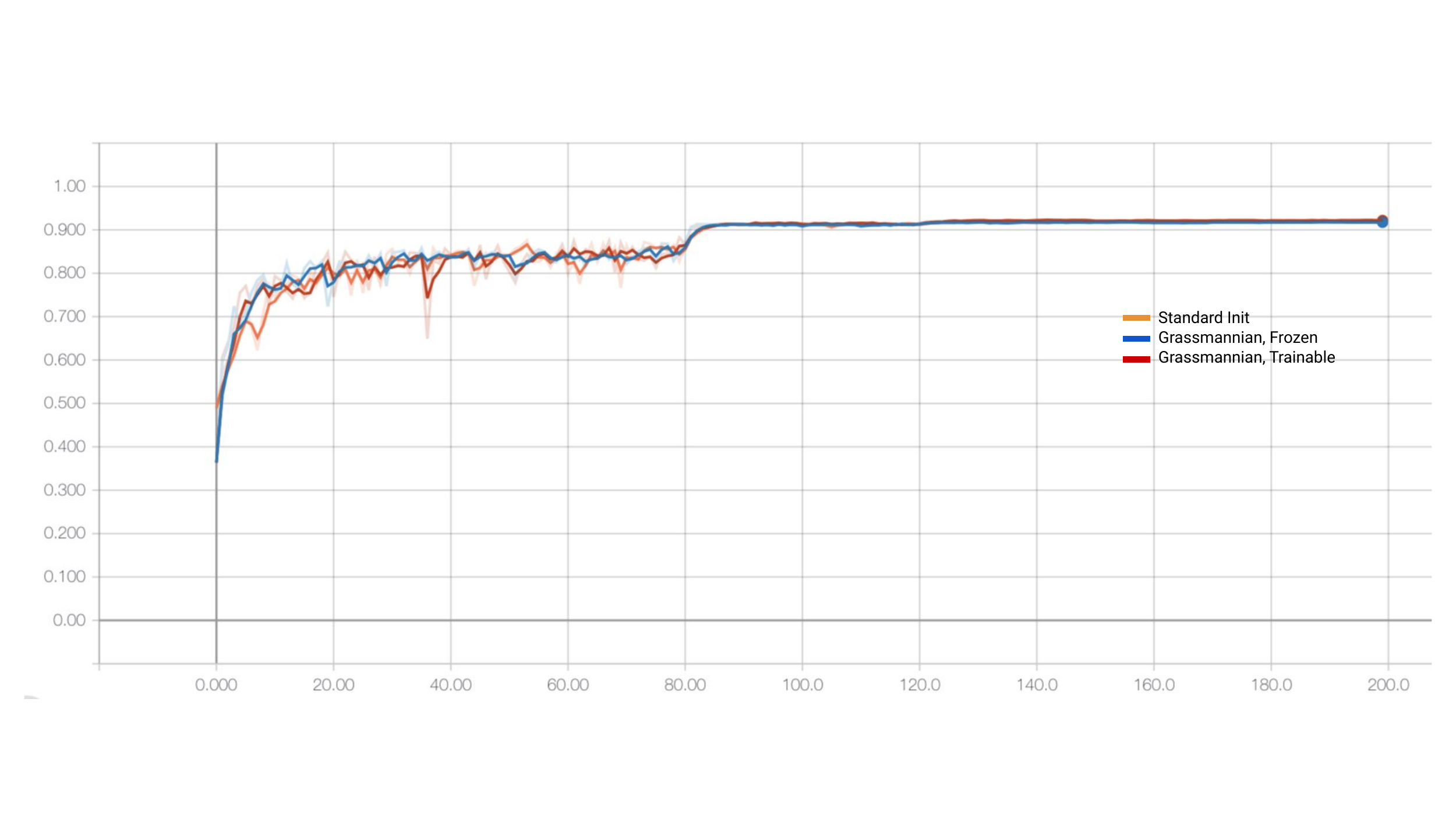}
    \caption{Comparison of standard initialization and Grassmannian initialization of first layer as both trainable and untrainable parameters on ResNet56 trained on CIFAR-10. Grassmannian approaches have a faster convergence with marginally better test accuracy with Adam optimizer used in all 3 cases.}
    \label{fig:resnet56}
\end{figure}

\end{document}